\documentclass[11pt]{article}

\usepackage[final]{acl}

\usepackage{times}
\usepackage{latexsym}

\usepackage[T1]{fontenc}

\usepackage[utf8]{inputenc}

\usepackage{microtype}

\usepackage{inconsolata}

\usepackage{graphicx}
\usepackage{amsmath}
\usepackage{amssymb}
\usepackage{graphicx}

\usepackage{multirow}
\usepackage{microtype}
\usepackage{natbib}
\usepackage{booktabs}
\usepackage{multirow}
\usepackage[table]{xcolor}
\usepackage{adjustbox}
\usepackage{booktabs}
\usepackage{multirow}
\usepackage{fontawesome5}
\usepackage{hyperref}
\definecolor{heatcyan}{RGB}{80,190,200}
\definecolor{heatorange}{RGB}{240,165,90}

\newcommand{\heatpos}[2]{\cellcolor{heatcyan!#1}#2}
\newcommand{\heatneg}[2]{\cellcolor{heatorange!#1}#2}
\title{Are You Sure You're Sure? On the Impact of Instruction Tuning on Confidence and Lexical Diversity}

\author{
  \textbf{Irina Proskurina\thanks{Equal contribution.}\textsuperscript{1,2}},
  \textbf{Mayank Kumar\footnotemark[1]\textsuperscript{1,3}},
  \textbf{Oyindolapo Komolafe\textsuperscript{1,4}}
  \\
  \textsuperscript{1}Cohere Labs Community
  \\
  \textsuperscript{2}Laboratoire Hubert Curien, UMR CNRS 5516, Saint-Étienne, France
  \\
  \textsuperscript{3}School of Computer Science Engineering and Technology (SCSET),\\
Bennett University, Greater Noida, India
  \\
  \textsuperscript{4}School of Physical Therapy, Faculty of Health Sciences, Western University, London, Canada
}

\begin{document}
\maketitle
\begin{abstract}

Instruction-tuned language models achieve strong performance across a range of generation tasks but have recently been shown to exhibit verbalized overconfidence, which may manifest in less diverse supporting rationales for incorrect answers. However, whether such overconfidence is associated with rationale consistency remains an open question.
In this paper, we study whether changes in the lexical diversity of generated answer rationales accompany changes in model confidence induced by instruction tuning.
We evaluate three matched base and instruction-tuned models across question-answering benchmarks and find that instruction tuning consistently increases answer confidence, despite limited changes in predictive accuracy, while degrading likelihood-based calibration. 
Secondly, we observe a non-uniform effect of instruction tuning on rationale diversity: cross-rationale diversity consistently decreases, whereas surface-level lexical diversity varies in both direction and magnitude across models and benchmarks. 
Finally, we find that these differences persist after controlling for answer selection and rationale length, confirming that confidence and rationale diversity capture distinct effects of instruction tuning.
\end{abstract}

\section{Introduction}\label{sec:intro}
Large language models (LLMs) are increasingly applied in question answering and reasoning tasks in medicine, finance, and law, where reliable estimates of model confidence are particularly important \citep{singhal2023large,hager2024evaluation,wu2023bloomberggpt,fei2024lawbench}.
Prior work has shown that training base models to follow natural-language instructions can improve model performance in such applications \citep{wei2021finetuned,ouyang2022training}.
At the same time, post-training was shown to alter model confidence distributions \citep{kadavath2022language,tian2023just,xiong2024can}.

\begin{figure}[t]
    \centering
    \includegraphics[width=0.95\linewidth]{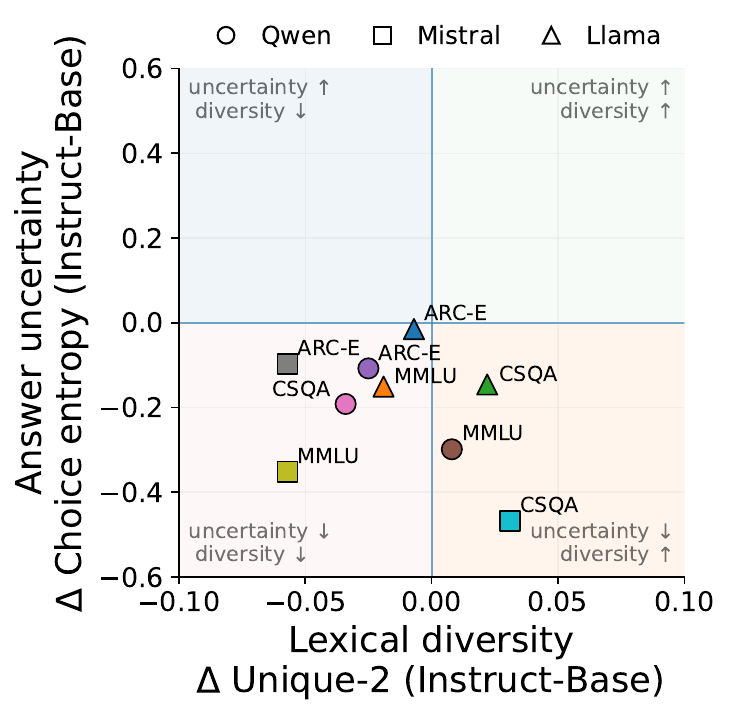}
    \caption{
    Effect of instruction tuning on answer uncertainty and lexical diversity of generated answer rationales.
    Each point represents a matched base-instruction model pair on a single benchmark, with score changes computed as $\Delta=\mathrm{Instruct}-\mathrm{Base}$.
    }
    \label{fig:intro_uncertainty_diversity}
\end{figure}

Model confidence, however, can be estimated using several proxies, including the probability assigned to the selected answer relative to alternative answers, consistency across repeated generations, and explicitly verbalized confidence \citep{kadavath2022language,tian2023just,xiong2024can}.
In free-form generation, uncertainty estimation is further complicated by variation in the surface form of generated responses, since semantically equivalent answers can be expressed using substantially different lexical sequences \citep{kapoor-etal-2024-calibration,farquhar2024detecting}.
While several recent studies investigate how instruction tuning and preference-based post-training affect model confidence \citep{zhang-etal-2024-calibrating,huang-etal-2026-investigating}, to the best of our knowledge, prior work has not examined whether changes in confidence are associated with changes in the lexical diversity of generated answer rationales, which provide supporting evidence for the selected answer.

In this work, we investigate how instruction tuning affects likelihood-based answer uncertainty and verbalized confidence, and whether the differences between base and instruction-tuned models are associated with corresponding changes in the lexical diversity of model rationales.

Our contributions are as follows: 1) we conduct a paired evaluation of base and instruction-tuned models across three model families and three reasoning benchmarks, comparing confidence, calibration, and rationale diversity; 2) we perform controlled comparisons restricted to examples for which base and instruction-tuned models select the same answer, while matching rationale length across variants, allowing us to measure changes in lexical diversity independently of answer switching and generation length; and 3) we show that instruction tuning increases model confidence without corresponding improvements in predictive accuracy, while changes in lexical diversity remain heterogeneous and are not consistently associated with uncertainty or calibration.
Together, our findings show that increased confidence after instruction tuning is not associated with a uniform shift in either rationale diversity or model calibration.
We make our code publicly available.\footnote{
\faGithub\ \href{https://github.com/upunaprosk/confidence-rationale-diversity}
{\texttt{upunaprosk/confidence-rationale-diversity}}
}
\section{Related Work}\label{sec:related-work}

Calibration measures how well a model's predicted confidence aligns with the empirical probability of being correct \citep{guo2017calibration,jiang-etal-2021-know,geng-etal-2024-survey}.
Several empirical studies have shown that model confidence can be misaligned with predictive accuracy, with over- and under-confidence observed for likelihood-based estimates \citep{kadavath2022language}, verbalized confidence \citep{tian2023just}, and generation-based uncertainty measures \citep{xiong2024can}.
Recent work has shown that post-training can further alter model confidence distributions, with alignment methods such as learning from human feedback affecting verbalized confidence even when downstream task performance improves \citep{zhu-etal-2023-calibration,tian2023just,zhang-etal-2024-calibrating,sanz-guerrero-etal-2026-large}.

Another line of work studies the impact of instruction tuning and other post-training on linguistic diversity in open-ended generation \citep{guo2025benchmarking,yun-etal-2025-price,deshpande-etal-2025-diverse}. 
Further works examine lexical, syntactic, and semantic diversity as indicators of creativity in model outputs, comparing the generated outputs with human-written text or human judgments of creativity \citep{tian-etal-2024-large-language,bae-kim-2024-collective,park-etal-2025-character,park-etal-2025-avoidance}.

To the best of our knowledge, despite a substantial body of work on output diversity, model confidence, and calibration, the relationship between lexical or token-level diversity and verbalized confidence or calibration has not yet been investigated.

\begin{table*}[t]
\centering
\resizebox{0.98\textwidth}{!}{
\begin{tabular}{llccccc|ccccc|ccccc}
\toprule
\multirow{2}{*}{\textbf{Model}} &
\multirow{2}{*}{} &
\multicolumn{5}{c}{\textbf{ARC-Easy}} &
\multicolumn{5}{c}{\textbf{MMLU}} &
\multicolumn{5}{c}{\textbf{CommonsenseQA}} \\

\cmidrule(lr){3-7}
\cmidrule(lr){8-12}
\cmidrule(lr){13-17}

&&
Acc.$\uparrow$
& $H_{\mathrm{choice}}\downarrow$
& Verb.$\uparrow$
& U2$\uparrow$
& $1$-SB$\uparrow$
& Acc.$\uparrow$
& $H_{\mathrm{choice}}\downarrow$
& Verb.$\uparrow$
& U2$\uparrow$
& $1$-SB$\uparrow$
& Acc.$\uparrow$
& $H_{\mathrm{choice}}\downarrow$
& Verb.$\uparrow$
& U2$\uparrow$
& $1$-SB$\uparrow$ \\

\midrule

\multirow{2}{*}{\textbf{Qwen2.5-7B}}
& Base
& \heatneg{41}{80.6}
& \heatneg{65}{.235}
& \heatneg{56}{49.6}
& \heatneg{1}{.694}
& \heatneg{7}{.695}
& \heatpos{61}{71.8}
& \heatpos{2}{.430}
& \heatneg{36}{59.8}
& \heatneg{41}{.679}
& \heatneg{24}{.671}
& \heatpos{47}{85.2}
& \heatpos{8}{.268}
& \heatneg{56}{27.4}
& \heatpos{65}{.751}
& \heatneg{15}{.734}
\\

& Instr.
& \heatneg{3}{81.6}
& \heatpos{54}{.127$^{*}$}
& \heatneg{26}{60.4$^{*}$}
& \heatneg{46}{.669$^{*}$}
& \heatneg{65}{.598$^{*}$}
& \heatpos{60}{71.7}
& \heatpos{65}{.131$^{*}$}
& \heatneg{14}{68.7$^{*}$}
& \heatneg{24}{.687}
& \heatneg{53}{.628$^{*}$}
& \heatpos{36}{82.5$^{*}$}
& \heatpos{38}{.076$^{*}$}
& \heatneg{34}{39.0$^{*}$}
& \heatneg{38}{.717$^{*}$}
& \heatneg{65}{.654$^{*}$}
\\

\midrule

\multirow{2}{*}{\textbf{Mistral-7B}}
& Base
& \heatneg{60}{80.1}
& \heatneg{43}{.215}
& \heatpos{18}{76.6}
& \heatpos{65}{.701}
& \heatpos{65}{.813}
& \heatneg{65}{59.6}
& \heatneg{51}{.680}
& \heatpos{27}{84.9}
& \heatpos{65}{.731}
& \heatpos{65}{.803}
& \heatneg{65}{57.4}
& \heatneg{65}{.736}
& \heatneg{18}{47.9}
& \heatneg{32}{.719}
& \heatpos{46}{.833}
\\

& Instr.
& \heatpos{65}{83.4$^{*}$}
& \heatpos{65}{.117$^{*}$}
& \heatpos{65}{93.8$^{*}$}
& \heatneg{37}{.673$^{*}$}
& \heatneg{48}{.626$^{*}$}
& \heatneg{64}{59.7}
& \heatpos{23}{.328$^{*}$}
& \heatpos{43}{91.5$^{*}$}
& \heatneg{51}{.674$^{*}$}
& \heatneg{55}{.626$^{*}$}
& \heatneg{17}{69.2$^{*}$}
& \heatpos{8}{.268$^{*}$}
& \heatpos{65}{92.2$^{*}$}
& \heatpos{62}{.750$^{*}$}
& \heatneg{9}{.745$^{*}$}
\\

\midrule

\multirow{2}{*}{\textbf{Llama-3.1-8B}}
& Base
& \heatpos{20}{82.2}
& \heatneg{13}{.188}
& \heatneg{57}{49.2}
& \heatpos{16}{.704}
& \heatpos{47}{.783}
& \heatneg{18}{64.1}
& \heatneg{36}{.608}
& \heatneg{65}{48.2}
& \heatpos{45}{.721}
& \heatpos{60}{.795}
& \heatneg{11}{70.8}
& \heatneg{6}{.362}
& \heatneg{20}{46.8}
& \heatneg{62}{.709}
& \heatpos{34}{.814}
\\

& Instr.
& \heatpos{20}{82.2}
& \heatpos{3}{.173$^{*}$}
& \heatpos{56}{90.4$^{*}$}
& \heatpos{4}{.697$^{*}$}
& \heatpos{9}{.720$^{*}$}
& \heatpos{26}{68.4$^{*}$}
& \heatneg{4}{.457$^{*}$}
& \heatpos{44}{91.8$^{*}$}
& \heatpos{6}{.702$^{*}$}
& \heatpos{8}{.719$^{*}$}
& \heatpos{10}{75.9$^{*}$}
& \heatpos{16}{.216$^{*}$}
& \heatpos{63}{91.1$^{*}$}
& \heatpos{5}{.731$^{*}$}
& \heatpos{9}{.773$^{*}$}
\\

\bottomrule
\end{tabular}
}
\caption{
Accuracy (Acc.), choice entropy $H_{\mathrm{choice}}$, verbalized confidence (Verb.), Unique-2 (U2), and $1$-$\mathrm{SelfBLEU}$ ($1$-SB) for paired base and instruction-tuned models across benchmarks.
Acc. and Verb. are reported in \%.
The colors indicate mean-centered values within each metric column (blue: above the mean; orange: below the mean).
${*}$ denotes a significant Base-Instruct difference at $p<0.01$ (two-sided paired $t$-test).
}
\label{tab:main_results}
\end{table*}

\section{Methodology}\label{sec:method}

We consider a multiple-choice question answering problem, where each input question $x$ is associated with a set of answers $Y={y_1,\ldots,y_M}$.

\paragraph{Model Confidence Evaluation}

Following \citealp{jiang-etal-2021-know}, we define the model prediction as the candidate answer with the highest conditional likelihood:
\begin{align}\label{eq:lik}
\hat{y} = \underset{y \in Y}{\arg\max}\; p_{\mathrm{LM}}(y\mid x).
\end{align}

We evaluate model confidence using several complementary measures. First, we estimate model uncertainty over the candidate answers using the entropy of normalized answer likelihoods \citep{malinin2020uncertainty,6773024}:
\begin{align}
H_{\mathrm{choice}}(x) = -
\frac{
\sum_{j=1}^{M}
p_j \log p_j
}{
\log M
},
\end{align}
where $p_j$ denotes the normalized probability assigned to candidate answer $y_j$ and higher values of $H_{\mathrm{choice}}(x)$ indicate greater uncertainty over the candidate answers.
Secondly, we also use elicited verbalized confidence as an estimate of model confidence, following \citet{xiong2024can}. Specifically, we use a two-stage protocol: in the first forward pass, we obtain the model prediction using the model likelihood in Eq.~\eqref{eq:lik}; then, in the second forward pass, we prompt the model to report a numerical probability that the selected answer is correct.
Additional implementation details are provided in \autoref{app:implementation-details}.

\paragraph{Calibration Evaluation}
Following \citet{guo2017calibration,jiang-etal-2021-know}, we compute Expected Calibration Error (ECE) as follows:
\begin{equation}
\mathrm{ECE}
=
\sum_{b=1}^{B}
\frac{|S_b|}{N}
\left|
\mathrm{acc}(S_b)-\mathrm{conf}(S_b)
\right|,
\end{equation}
where the $N$ predictions are partitioned into $B$ confidence bins $S_b$, $|S_b|$ denotes the number of predictions in bin $b$, $\mathrm{acc}(S_b)$ is the proportion of correct predictions in that bin, and $\mathrm{conf}(S_b)$ is the mean predicted confidence. 
We use $B=10$ equal-width confidence bins for both likelihood-based and verbalized confidence.
Lower ECE indicates better calibration.
We compute ECE for likelihood-based and verbalized confidence. 

\paragraph{Lexical Diversity}

Next, we measure lexical diversity in the rationales generated for each question and selected answer. For each question, we sample $K=5$ rationales using chain-of-thought prompting. The full prompt and generation details are provided in \autoref{app:implementation-details}.
We evaluate the generated rationales using the Unique Tokens Ratio and Self-BLEU measures, following prior work on diversity evaluation in text generation \citep{alihosseini-etal-2019-jointly}. In particular, we use the proportion of distinct bigrams as the Unique-2 score, with larger values indicating greater lexical richness. We also compute Self-BLEU as the similarity of each rationale to the remaining rationales generated for the same question, with larger values indicating greater similarity across generations. 

\paragraph{Experimental Settings}
We use three widely used multiple-choice benchmarks: ARC-Easy~\cite{clark2018think}, MMLU~\cite{hendrycks2020measuring}, and CommonsenseQA (CSQA)~\cite{talmor-etal-2019-commonsenseqa}. 
These benchmarks cover grade-school science, a broad range of academic and professional subjects, and commonsense knowledge grounded in ConceptNet \citep{speer2017conceptnet}, respectively.
For our analysis, we use three pairs of base and instruction-tuned models: Qwen2.5-7B, Llama-3.1-8B, and Mistral-7B-v0.3.
Model links and license information are provided in \autoref{app:implementation-details}.

\section{Results}\label{sec:results}
We begin by analyzing how instruction tuning affects the evaluated confidence and lexical diversity measures. The results across models and benchmarks are reported in \autoref{tab:main_results}.\footnote{We report $1$-$\mathrm{SelfBLEU}$ so that larger values correspond to greater cross-rationale variability.} We summarize the main findings below.
\paragraph{Instruction Tuning Consistently Increases Model Confidence without Corresponding Improvements in Accuracy.}
We find that across all models and benchmarks, instruction tuning consistently increases model confidence, as reflected by lower answer entropy and higher verbalized confidence (\autoref{tab:main_results}). Answer entropy decreases across all benchmarks, with particularly large decreases for Qwen on MMLU ($0.430$ to $0.131$) and Mistral on CSQA ($0.736$ to $0.268$). Similarly, verbalized confidence increases across all settings, including from $49.2\%$ to $90.4\%$ for Llama on ARC-Easy and from $46.8\%$ to $91.1\%$ on CSQA. In contrast, accuracy changes are less consistent across benchmarks. For example, on ARC-Easy, Llama accuracy remains unchanged at $82.2\%$, despite a significant increase in verbalized confidence and a decrease in choice entropy. 

\paragraph{Instruction Tuning Induces Heterogeneous Changes in Rationale Lexical Diversity.}
In contrast to model confidence, the effect of instruction tuning on lexical diversity is less consistent across models and benchmarks.
The largest decreases in Unique-2 are observed for Mistral on ARC-Easy and MMLU, while the largest increase occurs on CSQA, from $0.719$ to $0.750$.
At the same time, cross-rationale diversity, measured as $1$-$\mathrm{SelfBLEU}$, decreases across all models and benchmarks, with the largest decrease observed for Mistral on ARC-Easy, from $0.813$ to $0.626$.
Overall, instruction tuning consistently reduces cross-rationale variability, whereas changes in Unique-2 vary across models and benchmarks.

\paragraph{Decreases in Answer Uncertainty Do Not Consistently Coincide with Reduced Lexical Diversity.}
We further examine whether lower or higher answer uncertainty coincides with lower or higher lexical diversity. \autoref{tab:directional_csqa} reports the four possible directions of Instruct-Base uncertainty-diversity changes.
For the Qwen model, decreases in uncertainty most often coincide with decreases in diversity, accounting for $61.8\%$ of examples under Unique-2 and $69.3\%$ under $1-\mathrm{SelfBLEU}$. For Mistral, the pattern differs across diversity measures: lower uncertainty coincides with higher Unique-2 for $61.8\%$ of examples, but with lower $1-\mathrm{SelfBLEU}$ for $77.6\%$. Llama exhibits the same divergence between the two measures. Overall, decreases in answer uncertainty are not consistently accompanied by decreases in lexical diversity, and the direction of the association depends on the diversity measure and model. 
\autoref{fig:intro_uncertainty_diversity} illustrates
the same pattern, where choice entropy generally decreases after instruction tuning while lexical diversity changes in both directions across models and benchmarks.

\begin{table}[t]
\centering
\small
\resizebox{0.48\textwidth}{!}{
\begin{tabular}{llrrrr}
\toprule
\textbf{Model} & \textbf{Div.}
& $\downarrow\downarrow$
& $\downarrow\uparrow$
& $\uparrow\downarrow$
& $\uparrow\uparrow$ \\
\midrule

\multirow{2}{*}{Qwen2.5}
& U2    & \textbf{61.8} & 32.8 & 3.8 & 1.6 \\
& $1$-SB & \textbf{69.3} & 25.3 & 3.9 & 1.5 \\

\midrule

\multirow{2}{*}{Mistral}
& U2    & 36.0 & \textbf{61.8} & 0.5 & 1.7 \\
& $1$-SB & \textbf{77.6} & 20.2 & 1.3 & 0.9 \\

\midrule

\multirow{2}{*}{Llama-3.1}
& U2    & 33.3 & \textbf{47.5} & 7.2 & 12.0 \\
& $1$-SB & \textbf{54.7} & 26.1 & 12.4 & 6.8 \\

\bottomrule
\end{tabular}}

\caption{
Directional changes in choice entropy ($H$) and lexical diversity ($D$) for paired base and instruction-tuned models on CommonsenseQA.
The first and second arrows denote the Instruct-Base change in $H$ and $D$, respectively; e.g., $\downarrow\uparrow$ indicates lower uncertainty and higher diversity.
Values are reported as percentages of benchmark questions falling into each quadrant.
The most frequent directional pattern for each model and diversity measure is in bold.
}
\label{tab:directional_csqa}
\end{table}

\paragraph{Diversity Shifts Persist after Controlling for Answer Selection and Rationale Length.}
Because instruction tuning can affect both the selected answer and rationale length, we test whether the observed diversity changes persist when 1) matched base and instruction-tuned models select the same answer and 2) rationale length is matched across model variants.\footnote{For length matching, rationales are paired by length, and each pair is truncated to the length of the shorter rationale; see \autoref{app:implementation-details}.}
The evaluation results, together with the number of examples out of the 1,200 CSQA questions for which both conditions hold, are reported in \autoref{tab:controlled_csqa}.
We find that for Qwen, Unique-2 remains nearly unchanged ($-0.001$), while $1-\mathrm{SelfBLEU}$ decreases by $0.036$. For Mistral and Llama, Unique-2 increases significantly by $0.050$ and $0.053$, respectively, whereas $1-\mathrm{SelfBLEU}$ decreases by $0.069$ and $0.012$. Thus, the observed lexical diversity changes persist after controlling for answer selection and rationale length, while the two diversity measures continue to show different trends.

\begin{table}[t]
\centering
\small
\resizebox{0.48\textwidth}{!}{
\begin{tabular}{lrrr}
\toprule
\textbf{Model} & \textbf{$N$}
& $\Delta$U2
& $\Delta(1-\mathrm{SB})$ \\
\midrule
Qwen2.5-7B   & 1104 & $-.001$   & $-.036^{*}$ \\
Mistral-7B   & 904 & $+.050^{*}$ & $-.069^{*}$ \\
Llama-3.1-8B & 1065 & $+.053^{*}$ & $-.012^{*}$ \\
\bottomrule
\end{tabular}}

\caption{
Unique-2 (U2) and $1-\mathrm{SelfBLEU}$ ($1$-SB) changes for paired base and instruction-tuned models on CommonsenseQA, restricted to questions for which both variants select the same answer and have matched rationale lengths.
$^{*}$ denotes a statistically significant Base-Instruct difference at $p<0.01$ (two-sided paired $t$-test).
}
\label{tab:controlled_csqa}
\end{table}

\begin{table*}[h]
\centering
\resizebox{0.98\textwidth}{!}{
\begin{tabular}{llccc|ccc|ccc}
\toprule
\multirow{2}{*}{\textbf{Model}} &
\multirow{2}{*}{} &
\multicolumn{3}{c}{\textbf{ARC-Easy}} &
\multicolumn{3}{c}{\textbf{MMLU}} &
\multicolumn{3}{c}{\textbf{CommonsenseQA}} \\

\cmidrule(lr){3-5}
\cmidrule(lr){6-8}
\cmidrule(lr){9-11}

&&
Lik. ECE $\downarrow$
& Verb. ECE $\downarrow$
& Gap
& Lik. ECE $\downarrow$
& Verb. ECE $\downarrow$
& Gap
& Lik. ECE $\downarrow$
& Verb. ECE $\downarrow$
& Gap \\

\midrule

\multirow{2}{*}{\textbf{Qwen2.5-7B}}
& Base
& \textbf{6.9} & 35.3 & +28.4
& 4.4 & 40.8 & +36.4
& \textbf{1.6} & 59.1 & +57.5 \\

& Instruct
& 11.3 & 22.8 & +11.5
& 21.3 & 29.8 & \textbf{+8.5}
& 12.7 & 47.0 & +34.3 \\

\midrule

\multirow{2}{*}{\textbf{Mistral-7B}}
& Base
& 8.4 & 10.8 & +2.4
& 1.2 & 36.6 & +35.4
& 5.1 & 24.9 & +19.9 \\

& Instruct
& 10.2 & 11.9 & +1.7
& 22.3 & 35.4 & +13.1
& 15.2 & 27.6 & +12.4 \\

\midrule

\multirow{2}{*}{\textbf{Llama-3.1-8B}}
& Base
& 8.0 & 33.0 & +24.9
& \textbf{0.5} & \textbf{16.6} & +16.1
& 7.7 & 24.0 & +16.3 \\

& Instruct
& 8.5 & \textbf{8.5} & \textbf{0.0}
& 5.9 & 23.7 & +17.8
& 11.5 & \textbf{15.9} & \textbf{+4.5} \\

\bottomrule
\end{tabular}
}

\caption{
Expected Calibration Error (ECE) for paired base and instruction-tuned models across benchmarks.
Values are reported in \%.
The gap denotes the difference between verbalized confidence and likelihood-based ECE; negative values indicate lower calibration error for verbalized confidence than for likelihood-based confidence.
}

\label{tab:calibration-results-app}

\end{table*}

\paragraph{Changes in Lexical Diversity Are Not Consistently Associated with Calibration.}
We additionally compare changes in lexical diversity with model calibration, estimated using Expected Calibration Error (ECE). The full calibration results are reported in \autoref{tab:calibration-results-app}.
We find that changes in lexical diversity are not consistently associated with changes in calibration.
For instance, for the Qwen model, verbalized-confidence on ARC-Easy decreases from $35.3$ to $22.8$ after instruction tuning, while both Unique-2 and $1-\mathrm{SelfBLEU}$ decrease.
In contrast, for the Llama model on the MMLU benchmark, both likelihood-based and verbalized ECE increase ($0.5$ to $5.9$ and $16.6$ to $23.7$, respectively), while both diversity measures decrease.

Overall, the experimental results show that instruction tuning increases answer confidence while lexical diversity changes in different directions across models and measures, and that these confidence shifts are not consistently accompanied by improved calibration or predictable changes in rationale diversity.

\section{Conclusion}

In this paper, we investigate the impact of instruction tuning on model confidence and the lexical diversity of generated rationales in question answering tasks. 
Through our analysis of rationale diversity, we find that cross-rationale variability decreases after instruction tuning, whereas surface-level lexical diversity exhibits benchmark-dependent patterns. We further show that decreases in answer uncertainty are not consistently accompanied by decreases in lexical diversity, and that diversity changes do not consistently reflect changes in verbalized confidence, answer selection, or model calibration. Moreover, the same patterns persist after controlling for answer selection and rationale length.

Overall, our findings show that instruction tuning affects confidence and rationale diversity differently, motivating future work on uncertainty estimation that jointly considers predictive confidence and variation in generated rationales. Future work could further investigate whether reducing post-training overconfidence comes at the cost of further reducing rationale diversity, and whether the observed uncertainty-diversity patterns extend beyond lexical variation to semantic diversity.

\section*{Limitations}

To the best of our knowledge, this work represents one of the first attempts to jointly study rationale diversity and model verbalized and likelihood-based confidence. Our study is currently limited to three English multiple-choice benchmarks and lexical diversity measures. Future work may therefore extend the analysis to semantic and syntactic diversity \citep{guo2025benchmarking} of generated rationales across a broader range of benchmarks. Second, future work should examine whether the same uncertainty-diversity patterns generalize across languages.
More broadly, future work could examine how post-training calibration \citep{xie2024calibrating} or decoding interventions that modify generation probabilities \citep{meister-etal-2023-efficacy,nadeem-etal-2020-systematic} affect cross-rationale diversity in generation tasks.

\section*{Ethical Considerations}
We experiment with publicly available models and benchmark datasets and adhere to the intended use and licensing terms of the respective resources.

Our results nevertheless highlight potential risks associated with interpreting the confidence of instruction-tuned language models. Across the evaluated models and benchmarks, we find that instruction tuning increases model confidence and reduces cross-rationale diversity without a corresponding improvement in predictive accuracy.
In downstream applications, particularly in high-stakes settings, this mismatch may encourage unwarranted reliance on incorrect predictions when model confidence is interpreted as evidence that the prediction is reliable.
The differing calibration results obtained for likelihood-based and verbalized confidence further suggest that confidence should be assessed using multiple complementary measures rather than a single proxy, reducing the risk of inappropriate reliance on miscalibrated predictions.

Similarly, reduced cross-rationale diversity may limit the extent to which repeated generations provide independent evidence about a model's prediction. In generative settings, the implications of such shifts in rationale diversity should therefore be evaluated further, including comparisons against human-generated rationales and human judgments.

Finally, we do not evaluate instruction-tuned model performance on questions involving unsafe data or demographic-sensitive attributes. 
Consequently, our findings should not be interpreted as evidence that instruction tuning increases harmful or biased outputs for such questions.
Whether the observed confidence and diversity patterns extend to safety-sensitive or demographic-sensitive prompts requires separate evaluation.

\section*{Acknowledgments}

We thank Yanzhu Guo for insightful discussions and constructive feedback during the early stages of the project, Alvin Vinod Chandran, Abdelrahman Alkahwaji, and other members of the Cohere Labs Community for helpful exchanges and early feedback on the project statement.
We also thank Cohere Labs for providing collaborative opportunities throughout the development of this work.

This work was performed using HPC resources from GENCI-IDRIS (Grant 2025-AD011014384R1).

\bibliography{custom}

\newpage

\appendix
\onecolumn
\section{Examples of Generated Rationales}\label{app:rationales-examples}
\begin{table*}[h]
\centering
\footnotesize
\resizebox{0.98\textwidth}{!}{
\begin{tabular}{p{0.99\textwidth}}
\toprule
\textbf{Prompt.}
When a bottle of strong perfume is opened at the front of a classroom, the smell gradually spreads throughout the entire room. Which process explains this occurrence?
\textbf{A.} osmosis \quad
\textbf{B.} diffusion \quad
\textbf{C.} cell activity \quad
\textbf{D.} cell transport
\\
\texttt{Answer: Let's think step by step.}\\
\midrule
\textbf{Base}
\\
\textbf{R1.}
The perfume is most concentrated near the opened bottle and gradually spreads to other parts of the room. Diffusion is the movement of molecules from an area of higher concentration to an area of lower concentration. The perfume molecules therefore move away from the front of the classroom and mix with the surrounding air until they are more evenly distributed. Osmosis concerns water moving across a membrane, while cell activity and cell transport involve cellular processes. Therefore, the correct answer is B. diffusion.
\\
\textbf{R2.}
First, the smell starts near the bottle, where the concentration of perfume molecules is highest. Over time, those molecules spread through the air toward regions where their concentration is lower. This is the definition of diffusion. Osmosis requires the movement of water across a selectively permeable membrane, which is not involved here. Cell activity and cell transport also do not describe movement through open air. Therefore, the process responsible for the spreading perfume smell is B. diffusion.
\\
\textbf{R3.}
The perfume is released at the front of the classroom and then spreads through the surrounding air. The air initially contains many perfume molecules near the bottle and fewer molecules farther away. Random molecular motion produces a net movement from the region of higher concentration to regions of lower concentration. That process is diffusion. No cell or membrane is required, so osmosis, cell activity, and cell transport are not appropriate explanations. The correct answer is therefore B. diffusion.
\\
\textbf{R4.}
I can solve this by eliminating the alternatives. Cell activity and cell transport describe processes involving cells, whereas the perfume is spreading through the air. Osmosis specifically refers to the movement of water through a selectively permeable membrane, so it also does not fit. Diffusion describes particles spreading from a region of higher concentration to a region of lower concentration. Since the perfume molecules spread from the bottle throughout the classroom, the process is diffusion. The answer is B.
\\
\textbf{R5.}
When the bottle is opened, perfume molecules enter the air at a high concentration near the front of the room. Because molecules move randomly, they gradually spread into regions where fewer perfume molecules are present. This continues until the molecules become more evenly distributed throughout the room. Such movement from higher to lower concentration is diffusion. Osmosis is limited to solvent movement across a membrane, and the cellular options are irrelevant here. Thus, B. diffusion is the best answer.
\\

\midrule
\textbf{Instruct}

\textbf{R1.}
The perfume molecules are initially concentrated near the opened bottle. After release, they spread through the classroom from this region of high concentration toward regions of lower concentration. This movement continues as the molecules mix with the surrounding air and become more evenly distributed. The process is diffusion. Osmosis involves water moving across a selectively permeable membrane, while cell activity and cell transport concern processes within cells. Therefore, the correct answer is B. diffusion.
\\
\textbf{R2.}
Opening the perfume bottle releases perfume molecules into the surrounding air. These molecules move randomly and spread from the area near the bottle, where their concentration is high, toward areas of the classroom where their concentration is lower. This process is known as diffusion. Osmosis concerns the movement of water across a semi-permeable membrane, and the two cellular options do not apply to particles spreading through air. Therefore, the correct answer is B. diffusion.
\\
\textbf{R3.}
The smell spreads because perfume molecules move through the air from the point where they are released to the rest of the classroom. Near the bottle, the concentration of perfume molecules is initially higher than it is farther away. Their net movement toward lower-concentration regions is called diffusion. Osmosis requires a membrane and specifically concerns solvent movement, while cell activity and cell transport involve cellular mechanisms. Hence, the process described in the question is B. diffusion.
\\
\textbf{R4.}
The key observation is that the perfume spreads from a concentrated source into the larger volume of air in the classroom. Particles naturally move and mix, producing a net movement from regions of higher concentration to regions of lower concentration. This is diffusion. The process does not require a cell or a membrane, so cell activity, cell transport, and osmosis do not explain the observation. Therefore, the smell spreads throughout the room by diffusion, corresponding to answer B.
\\
\textbf{R5.}
Perfume molecules are released from the bottle into the air, creating a high concentration near the source. As the molecules move, they spread toward parts of the room where the concentration is lower and gradually become distributed throughout the available space. This is the characteristic process of diffusion. Osmosis is a specific membrane-related process involving solvent molecules, and cell activity or cell transport is not relevant in this setting. Thus, the correct answer is B. diffusion.
\\

\bottomrule
\end{tabular}}

\caption{
Example rationales generated by the Base and Instruct variants of Mistral-7B for the same ARC-Easy question.
}
\label{tab:qualitative_rationales}

\end{table*}

\newpage
\twocolumn
\section{Implementation Details}\label{app:implementation-details}

\paragraph{Rationale Generation}
To estimate variability under repeated generation, we sample $K=5$ rationales for each question across the considered benchmarks. Following the Zero-shot Chain-of-Thought prompting approach introduced by \citet{kojima2022large}, we use the prompt \texttt{``Answer: Let's think step by step.''} after each question and its candidate answers.
Rationales are sampled with temperature $T=0.7$, nucleus sampling $p=1.0$, and a maximum of 100 newly generated tokens. We generate five rationales per question. The generation settings are fixed across models and benchmarks. 
We provide a few examples of generated rationales in \autoref{tab:qualitative_rationales}.

\paragraph{Controlled Lexical Analysis}
A direct comparison of rationale diversity between Base and Instruct models can be affected by differences in their selected answers or generated rationale lengths. We therefore conduct an additional controlled analysis discussed in \S\ref{sec:results} (\autoref{tab:controlled_csqa}) restricted to examples for which the Base and Instruct variants select the same answer. 
Within each example, we retain rationales supporting this common selected answer, match the number of rationales between the two variants, and match rationale length by pairing rationales according to token length and truncating each pair to the length of the shorter rationale before computing lexical-diversity measures. This analysis isolates differences in the linguistic diversity of a fixed answer from differences caused by answer selection or unequal generation length.

\paragraph{Likelihood-Based Confidence and Statistical Analysis}
We use the LM Evaluation Harness \citep{biderman2024lessons} for benchmark accuracy evaluation.
For each multiple-choice question, we extract the normalized log-likelihood score $\ell_j$ for each candidate answer from the LM Evaluation Harness. We apply a softmax to the scores to obtain the answer probabilities $p_j$, which are used to compute likelihood-based confidence and choice entropy.

Self-BLEU is computed with SacreBLEU \citep{post-2018-call}.
We use two-sided paired $t$-tests on the per-example Instruct-Base differences.
We denote differences with $^{*}$ when $p<0.01$.
Statistical tests for the controlled lexical analysis are restricted to examples satisfying the same-answer, matched-rationale-count, and rationale-length conditions described above. 
Unique-$n$ is computed jointly across the five generated rationales for each question.

\paragraph{Verbalized Confidence Evaluation}
We obtain verbalized confidence using a separate two-stage prompting procedure, following \citet{xiong2024can}. 
The model's answer is first fixed to the answer selected by likelihood-based multiple-choice evaluation. We then prompt the model to estimate the probability that this selected answer is correct. Conditioning on the previously selected answer prevents the verbal-confidence stage from introducing a second, potentially different prediction.
The model output is constrained to a numerical probability. 
We verify a 100\% parsing rate across all model variants and benchmarks.
We provide below the verbalized-confidence prompt.
\begin{quote}
\small
\texttt{Question: [question]}\\
\texttt{A. [choice A]}\\
\texttt{B. [choice B]}\\
\texttt{\ldots}\\[2pt]
\texttt{Selected answer: [selected answer]}\\
\texttt{What is the probability that the selected answer is correct?}\\
\texttt{Give only a number between 0 and 1.}
\end{quote}

\paragraph{Model Links.}
Model references and license information are provided in \autoref{tab:model-details}.
All models used in this study are publicly available, and the usage of these models in our experiments complies with the intended use and applicable licensing terms of the respective resources.
\begin{table*}[h]
\centering
\footnotesize
\begin{tabular}{lll}
\toprule
\textbf{Model} & \textbf{Link} & \textbf{License} \\
\midrule
Qwen2.5-7B &
\url{https://hf.co/Qwen/Qwen2.5-7B} &
Apache 2.0 \\
&
\url{https://hf.co/Qwen/Qwen2.5-7B-Instruct} &
Apache 2.0 \\
\midrule
Llama-3.1-8B &
\url{https://hf.co/meta-llama/Llama-3.1-8B} &
Llama 3.1 Community License \\
&
\url{https://hf.co/meta-llama/Llama-3.1-8B-Instruct} &
Llama 3.1 Community License \\
\midrule
Mistral-7B-v0.3 &
\url{https://hf.co/mistralai/Mistral-7B-v0.3} &
Apache 2.0 \\
&
\url{https://hf.co/mistralai/Mistral-7B-Instruct-v0.3} &
Apache 2.0 \\
\bottomrule
\end{tabular}
\caption{Models used in the experiments with the associated licenses.}
\label{tab:model-details}
\end{table*}

\section{Extended Results}
\label{app:extended-evaluation-results}
In this appendix, we report additional analyses that extend the main results across benchmarks and provide further analysis of uncertainty, lexical diversity, and rationale length.
\paragraph{Uncertainty-Diversity Directional Analysis}
\autoref{tab:directional_appendix} extends the question-level directional analysis reported in \autoref{tab:directional_csqa} to ARC-Easy and MMLU. For each question, we report the direction of the Instruct-Base change in choice entropy ($H$) together with the corresponding change in lexical diversity ($D$). The results further show that decreases in answer uncertainty can coincide with either increases or decreases in lexical diversity, depending on the model and diversity measure.

\begin{figure*}[h]
    \centering
    \includegraphics[width=0.56\linewidth]{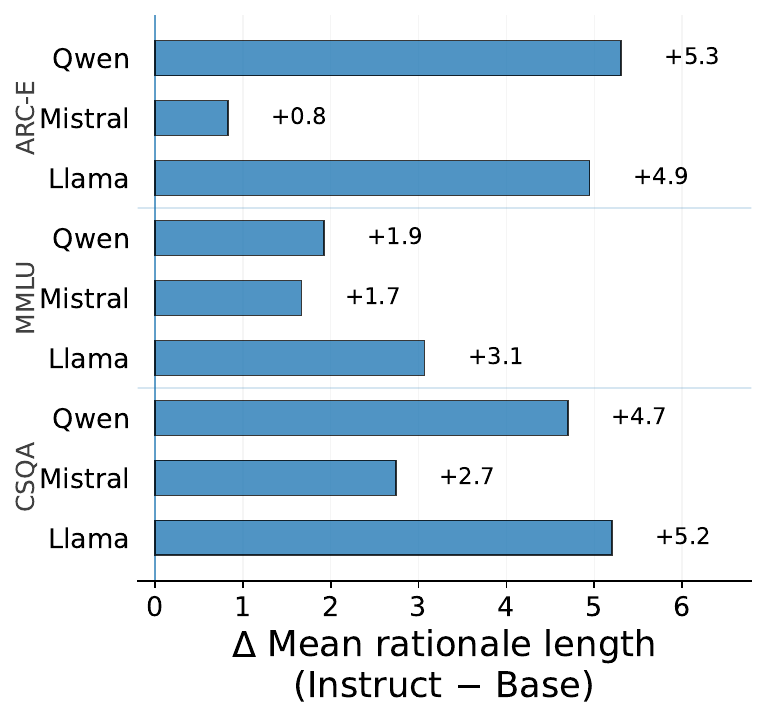}
    \caption{
Average change in rationale length from Base to Instruct models across benchmarks.
Positive values indicate longer rationales after instruction tuning, whereas negative values indicate shorter rationales.
}\label{fig:rationale_length}
\end{figure*}

\paragraph{Rationale Length Evaluation}
\autoref{fig:rationale_length} reports changes in mean rationale length between Base and Instruct models across benchmarks. Rationale length increases across all models and benchmarks, motivating the length-controlled analysis in \autoref{tab:controlled_csqa}.

\begin{table*}[h]
\centering
\small
\setlength{\tabcolsep}{4.5pt}
\renewcommand{\arraystretch}{1.08}

\begin{tabular}{lllrrrr}
\toprule
\textbf{Dataset}
& \textbf{Model}
& \textbf{Diversity}
& $H\downarrow D\downarrow$
& $H\downarrow D\uparrow$
& $H\uparrow D\downarrow$
& $H\uparrow D\uparrow$ \\
\midrule

\multirow{6}{*}{ARC-Easy}

& \multirow{2}{*}{Qwen2.5-7B}
& U2    & \textbf{52.4} & 34.7 & 7.7 & 5.2 \\
&
& $1$-SB & \textbf{65.7} & 21.4 & 9.3 & 3.6 \\

\cmidrule(lr){2-7}

& \multirow{2}{*}{Mistral-7B}
& U2    & \textbf{64.3} & 25.5 & 7.6 & 2.6 \\
&
& $1$-SB & \textbf{82.6} & 7.2 & 9.5 & 0.7 \\

\cmidrule(lr){2-7}

& \multirow{2}{*}{Llama-3.1-8B}
& U2    & \textbf{32.2} & 29.8 & 20.4 & 17.6 \\
&
& $1$-SB & \textbf{45.5} & 16.5 & 26.2 & 11.8 \\

\midrule

\multirow{6}{*}{MMLU}

& \multirow{2}{*}{Qwen2.5-7B}
& U2    & 48.1 & \textbf{50.1} & 0.6 & 1.2 \\
&
& $1$-SB & \textbf{62.2} & 36.0 & 1.0 & 0.8 \\

\cmidrule(lr){2-7}

& \multirow{2}{*}{Mistral-7B}
& U2    & \textbf{68.7} & 28.9 & 1.7 & 0.7 \\
&
& $1$-SB & \textbf{86.6} & 11.0 & 2.1 & 0.3 \\

\cmidrule(lr){2-7}

& \multirow{2}{*}{Llama-3.1-8B}
& U2    & \textbf{50.4} & 33.8 & 8.8 & 7.0 \\
&
& $1$-SB & \textbf{62.2} & 22.0 & 12.1 & 3.7 \\

\bottomrule
\end{tabular}

\caption{
Directional changes in choice entropy ($H$) and lexical diversity ($D$) for paired base and instruction-tuned models on ARC-Easy and MMLU benchmarks.
The first and second arrows denote the Instruct-Base change in $H$ and $D$, respectively; e.g., $\downarrow\uparrow$ indicates lower uncertainty and higher diversity.
Values are reported as percentages of benchmark questions falling into each quadrant.
The most frequent directional pattern for each model and diversity measure is in bold.
}
\label{tab:directional_appendix}
\end{table*}

\end{document}